# A Robust Rapid Approach to Image Segmentation with Optimal Thresholding and Watershed Transform

Ankit Chadha, Neha Satam

Vidyalankar Institute of Technology, Mumbai

## ABSTRACT

This paper describes a novel method for partitioning image into meaningful segments. The proposed method employs watershed transform, a well-known image segmentation technique. Along with that, it uses various auxiliary schemes such as Binary Gradient Masking, dilation which segment the image in proper way. The algorithm proposed in this paper considers all these methods in effective way and takes little time. It is organized in such a manner so that it operates on input image adaptively. Its robustness and efficiency makes it more convenient and suitable for all types of images.

## General Terms

Image Processing

## 1. INTRODUCTION

Due to advent of various techniques, image processing has been an attractive topic amongst enthusiasts. But it is equally challenging as an image represents various shapes, colors, tonal gradations, intensities and this information should be preserved in processing. It largely depends on features extracted, their uniqueness and their correctness. Though the features can vary from image to image, there are few common features, such as edge or boundary between object and background. Segmentation is one useful method for processing such images.

Image segmentation refers to process of partitioning an image into groups of pixels which are homogeneous with respect to some criterion [1]. It separates out foreground and background in meaningful way. It basically detects the edge of objects by finding transitions in brightness of image. Although segmentation by edge detection is one of the major tools to aid range image analysis, it is still considered an unsolved problem [2]. An algorithm working for some type of images may not work for some other. Hence it should be adaptive in nature. Following should be some properties followed by any algorithm:

1.  Universality: the algorithm should work for all types of images. It should be adaptive. Authors in [3]-[5] have developed a method for such adaptive image segmentation.

2.  Robust: algorithm should be robust in every aspect.

3.  Performance: in order to implement it practically, method should be highly efficient and yielding best outputs. It should achieve accuracy and give desired performance.

4.  Time efficient: it should achieve high speed to provide best outputs.

Segmentation can also be thought as pixel labeling process where the pixels belonging to the same homogeneous region are assigned the same label. There are many eigenvector-based methods for image segmentation but they are too slow for practical applications [6]. For error-free and rapid operation, some of the techniques can be given as below [7]:

1.  Histogram-based segmentation: If the image is composed of regions with different grey level ranges, i.e., the regions are distinct, the histograms of image usually shows different peaks, each corresponding to one region and adjacent peaks are likely to be separated by a valley [8]. Thus, when histogram has valleys, selection of threshold becomes easy because it becomes a problem of detecting valleys. Though this is an easy method, it has its own drawbacks. As only histogram information is taken into consideration, such algorithm may fail to detect thresholds if they are not properly reflected as valleys in histogram.

2.  Edge-based segmentation: In edge-based methods, the local discontinuities are detected first and then connected to form longer, complete boundaries [9]. It is the most common approach for detecting meaningful discontinuities in the grey level. An edge is a vector variable with two components magnitude and orientation, where edge magnitude gives the amount of the difference between pixels in the neighborhood (the strength of the edge) and edge orientation gives the direction of the greatest change, which presumably is the direction across the edge [10]. But it becomes difficult for noisy images.

3.  Region-based segmentation: In region-based methods, areas of an image with homogeneous properties are found, which in turn give the boundaries [9]. It relies on common patterns in intensity values within a cluster of neighboring pixels. The cluster is referred to as the region. Here the image is first segmented into predefined small regions, and then each predefined region is assigned a single motion label [11]. This region-based label assignment approach facilitates obtaining spatially continuous segmentation maps which are closely related to actual object boundaries.

4.  Hybrid-techniques segmentation: In this technique, various methods above described are combined. They are mainly centered on edge-based and region-based techniques. For many times, edge is defined initially and then image is segmented into different regions.

The algorithm proposed here belongs to the fourth category, i.e., hybrid-based technique. This paper tries to compare segmentation with and without thresholding, where the method used for segmentation is watershed segmentation, a well-known segmentation method which considers the image to be processed as a topographic surface. The topographic surface is most often built from an image gradient, since object edges (also known as watershed lines) are most probably located at pixels with high gradient values. It is a simple method which is less time-consuming and provides closest contours even for the regions having low contrast and weak boundaries.

The paper is organized as follows: section 2 gives brief idea about watershed transform and watershed segmentation. Section 3 depicts steps of implementation. Experimental results are provided in Section 4 and conclusions are presented in Section 5.

## 2. WATERSHED TRANSFORM AND WATERSHED SEGMENTATION

Concept of watershed is well known in topography. A watershed is a basin-like landform defined by highpoints and ridgelines that descend into lower elevations and stream valleys. It separates out different regions depending on their depth.

In image processing, a grayscale image is considered to be a topographic region. Every gray level denotes altitude of that region. If the surface is flooded from minima and merging of waters is prevented then image is partitioned into two distinct regions- catchment basin and watershed lines. Figure 1 depicts pictorial representation of these phrases [12].

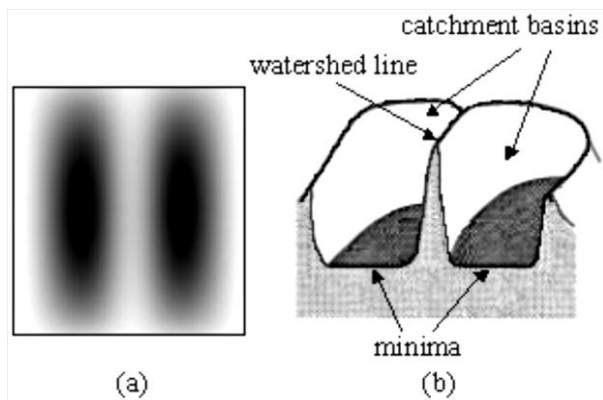

**Figure 1: Pictorial representation of watershed transform**

Catchment basin is set of those points at which a drop of water will certainly fall to a single minimum whereas watershed line is set of those points at which a drop of water will be equally likely to fall to more than one minimum. They are generally crest lines on the topographic surface. Aim of watershed transform is to find all the watershed lines.

Watershed can be defined for two cases-continuous and discrete.

### 2.1 Continuous

Assume that the image f is an element of the space C (D) of real twice continuouslydifferentiable functions on a connected domain D with only isolated critical points. Then the topographical distance between points p and q in D is defined by:

$$Tf(p,q) = \inf_{\gamma} \int_{\gamma} \| \nabla f(\gamma(s)) \| \, ds \qquad (1)$$

where the infimum is over all paths (smooth curves) γ inside D with γ (0) = p, γ (1) = q.

The topographical distance between a point p ∈ D and a set A ⊆ D is defined as:

Tf (p,A) =MIN$_a$ ⊆ $_A$ Tf (p, a).     (2)

The path with shortest Tf -distance between p and q is a path of steepest slope.

### 2.2 Discrete

A problem which arises for digital images is the occurrence of plateaus, i.e., regions of constant grey value, which may extend over large image areas. Such plateaus form a difficulty when trying to extend the continuous watershed definition based on topographical distances to discrete images.

Thus watershed transform can be defined as follows [13]:

Let f ∈ C(D) have minima {mk}k ∈ I, for some index set I. the catchment basin CB(mi) is defined as the set of points x ∈ D which are topographically closer to mi than any other regional minimum mj :

CB(m$_i$)=     {x ∈ D | ∀j∈ I{i}: f(mi) + Tf(x,m$_i$)< f(m$_j$) + Tf(x,m$_j$) }
(3)

The watershed of f is the set of points which do not belong to any catchment basin:

Wshed (f) = D ∩ (∪$_{i∈I}$ CB(mi))$^c$
(4)

Let W be some label where W ∉ I. the watershed transform of f is a mapping λ :D→ I U {W}, such that λ(p)= I if p ∈ CB(mi) and λ(p)= W if p ∈Wshed (f).

So the watershed transform of f assigns labels to the points of D, such that

(i)     Different catchment basins are uniquely labeled, and

(ii)    A special label W is assigned to all points of the watershed of f.

## 3. STEPS OF IMPLEMENTATION

Considering an image as input, all the steps were carried out. The image can be colored or gray; the algorithm works properly on both types of images. Flowcharts for both-with and without thresholding are shown in Figure 2.

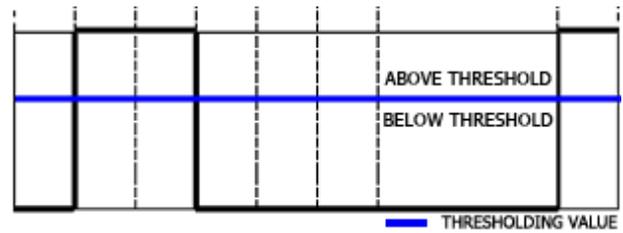

**Figure 3: Thresholding operation**

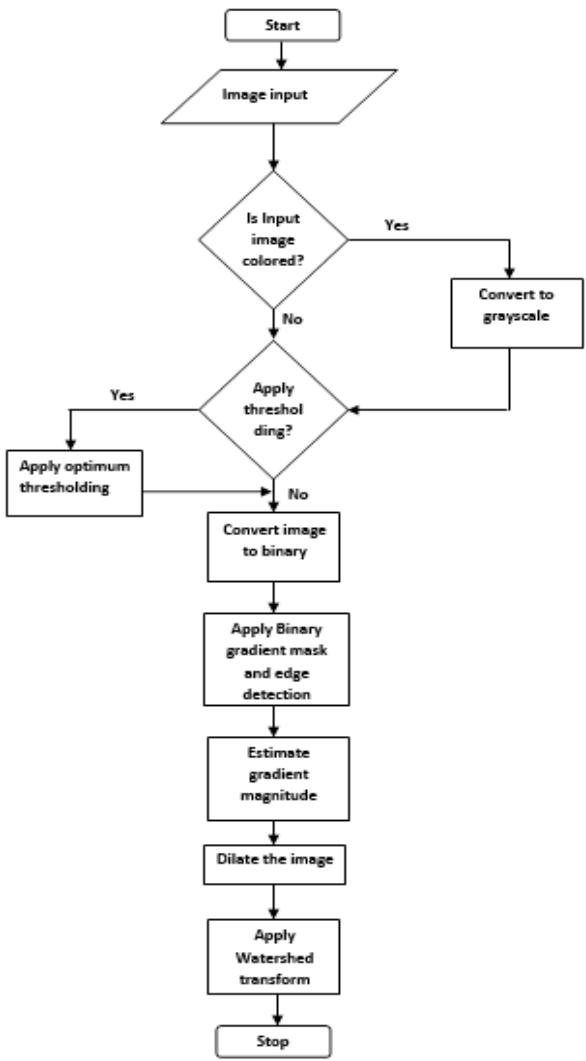

**Figure 2: Flowchart of algorithm**

## 3.1 Thresholding

The image taken is firstly converted to grayscale if it is colored. Then thresholding is applied on gray image. It is simplest kind of image segmentation. It separates out regions of an image corresponding to objects which we want to analyze. This separation is based on the variation of intensity between the object pixels and the background pixels. To differentiate the pixels a comparison of each pixel intensity value is performed with respect to a threshold, known as thresholding value. The thresholding value is determined from the intensity of image in grayscale.

The thresholding operation can be mathematically explained as follows:

I(x,y) = max_value if src(x,y) >thresholding_value

= 0                     otherwise

Figure 3 shows operation of thresholding.

Thus if the intensity of the pixel src(x,y) is higher than thresholding_value, then the new pixel intensity is set to a max_value.Otherwise, the pixels are set to 0.

In this paper, we have used optimal thresholding so as to perform the task for every possible image. It acts as an improvement and systematizes the process of manual thresholding.

## 3.2 Binary gradient masking [14]:

When considering a grayscale image, the objects to be detected vary greatly in contrast from their background. Changes in contrast can usually be distinguished by particular operators that calculate the gradient of an image. During calculating the gradient of that image thresholding value is first determined and then applied in subsequent steps to create a binary gradient mask (BGM). The threshold value will be tuned for second use under another combination of the same operator in order to obtain the required BGM image containing the segmented objects.

## 3.3 Dilation:

The dilation operator takes two pieces of data as input: a binary image, which is to be dilated and a structuring element (or kernel), which determines the behavior of the morphological operation.

To compute the dilation of a binary input image by this structuring element, we consider each of the background pixels in the input image in turn. For each background pixel (also known as input pixel) the structuring element is superimposed on top of the input image so that the origin of the structuring element coincides with the input pixel position. If at least one pixel in the structuring element coincides with a foreground pixel in the image underneath, then the input pixel is set to the foreground value. If all the corresponding pixels in the image are background, however, the input pixel is left at the background value.

Considering that X is the set of Euclidean coordinates of the input image, and K is the set of coordinates of the structuring element let Kx denote the translation of K so that its origin is at x. The dilation of X by K is simply the set of all points x such that the intersection of Kx with X is non-empty. Thus

$$\delta_B A = A \oplus B \bigcup_{a \in A} B_\alpha$$

dilation of set A by structuring element B is defined as:      (5)

Suppose that the structuring element is a 3x3 square with the origin at its center as shown in Figure 4 (a).

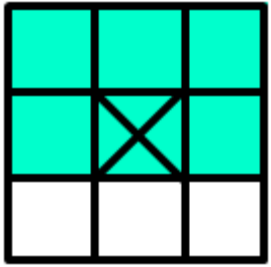

**Figure 4 (a): Structuring element**

And let the input image is as shown in Figure 4(b).

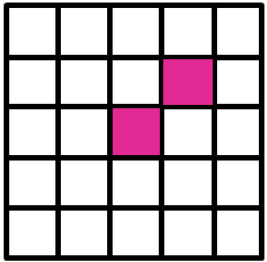

**Figure 4(b): Input image**

Then after dilating, we get Figure 4(c) as output.

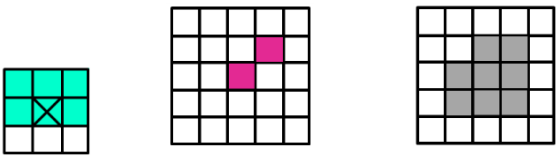

**Figure 4(c): Output of dilation**

### 3.4 Watershed transform

As explained in section II, watershed transform is based on flooding simulation [15]. Following are the steps included in this:

1. Piercing holes in each regional minimum of image

2. The 3D topography is flooded from below gradually

3. When the rising water in distinct catchment basins is about to merge, a dam is built to prevent the merging

Figure 5 depicts the flooding process on one-dimensional signal with four regional minima generating four catchment basins [15].

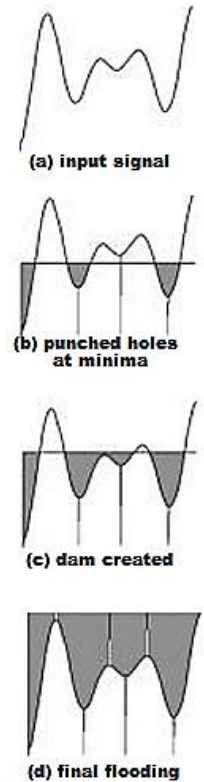

**Figure 5: Flooding process**

Final flooding shows three watershed lines and four catchment basins.

## 4. EXPERIMENTAL RESULTS

The 2-D (768x1024, 8 b/pixel) image shown in Figure 6(a) was used in order to illustrate the stages of the segmentation algorithm and visually assess the quality of the segmentation results. Figure 6(b) is grayscale version of original image. Figure 7(a) shows a binary conversion of image 6(b) which was carried out for segmentation without thersholding. On the binary image, binary gradient mask was applied, whose result is as shown in Figure 7(b). Using the Sobel edge masks, gradient magnitude was calculated in Figure 7(c). The gradient is supposed to be high at the borders of the objects and low inside the objects. Figure 7(d) shows the result of the dilation. Watershed transform is taken in Figure 7(e).

For segmentation using thresholding, Figure 8(a) is very useful as it shows thresholded image. On this thresholded image, BGM was applied and result is depicted in Figure 8(b). The gradient magnitude was calculated in Figure8(c). In Figure 8(d), dilated image is shown, on which watershed transform was applied. This result is shown in Figure 8(e).

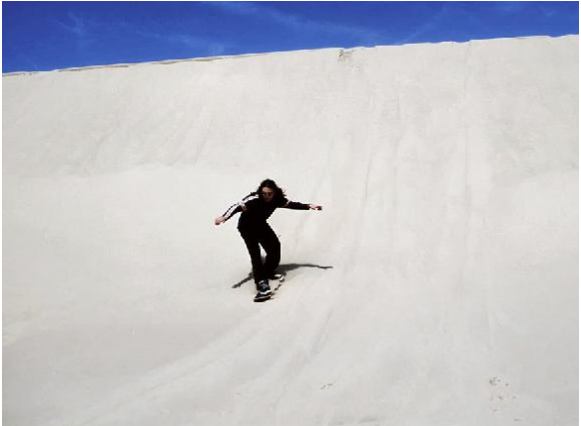

**Figure 6 (a): Original Image**

**Without Threshold**

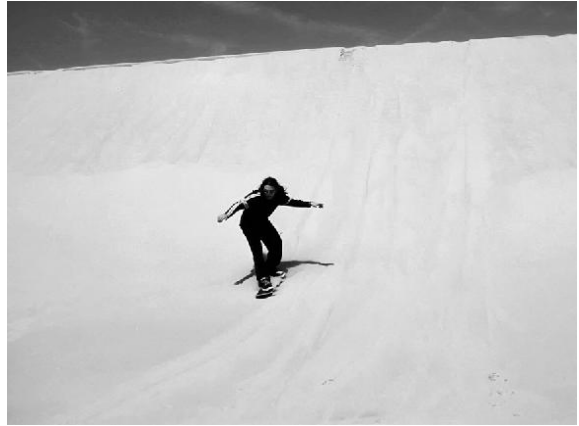

**Figure 6(b): Grey-Scaled Image**

**With Threshold**

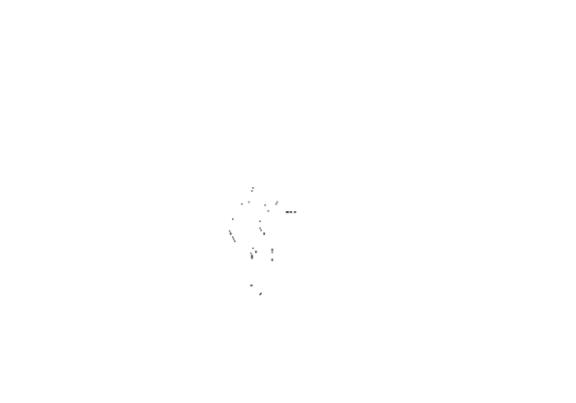

**Figure 7 (a): binary image**

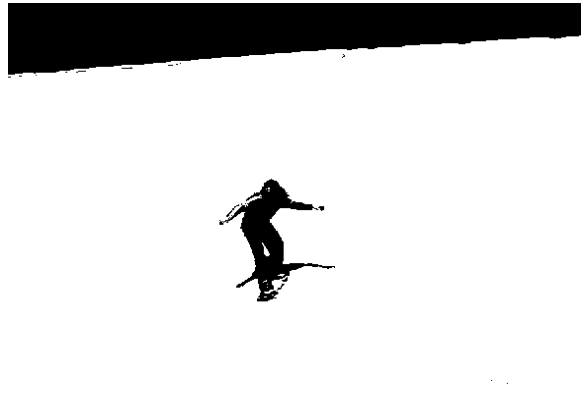

**Figure 8 (a): Thresholded image**

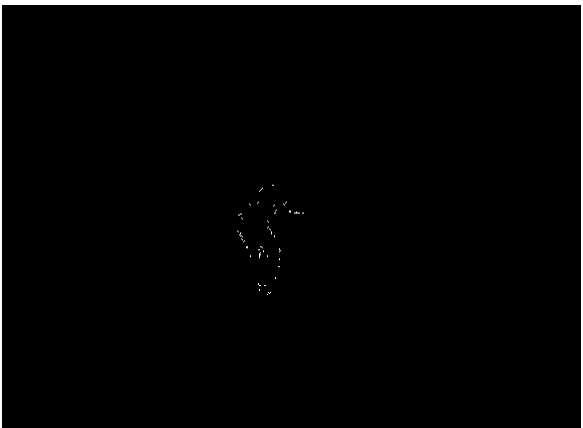

**Figure 7 (b): Binary Gradient Masked Image**

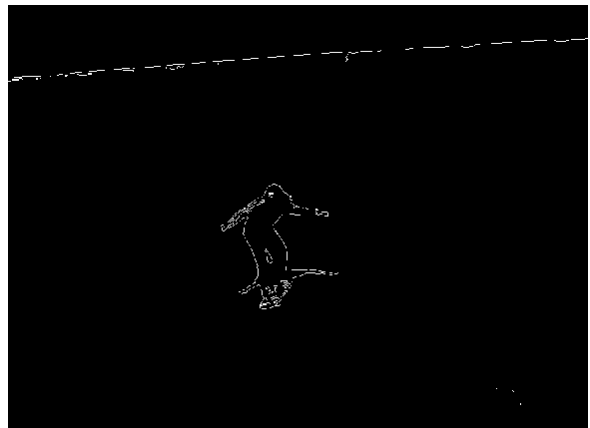

**Figure 8 (b): Binary Gradient Masked Image**

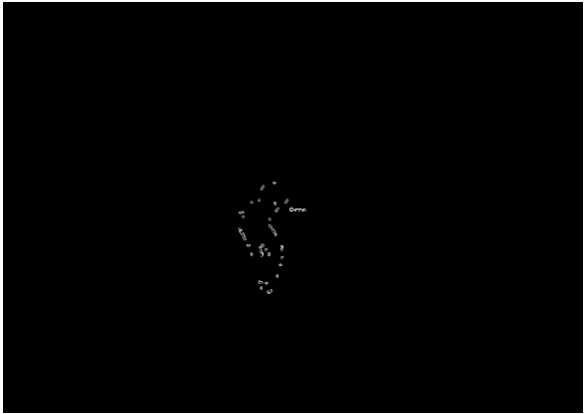

**Figure 7 (c): Gradient Magnitude**

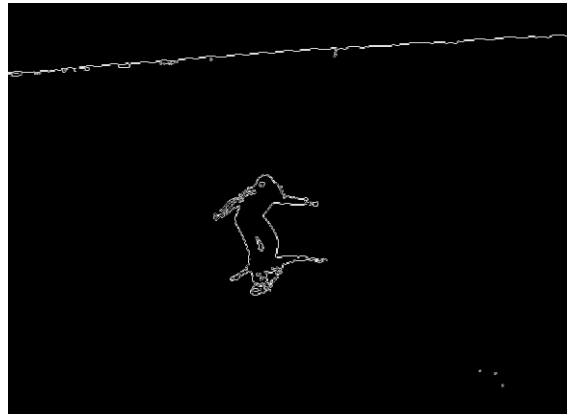

**Figure 8 (c): Gradient Magnitude**

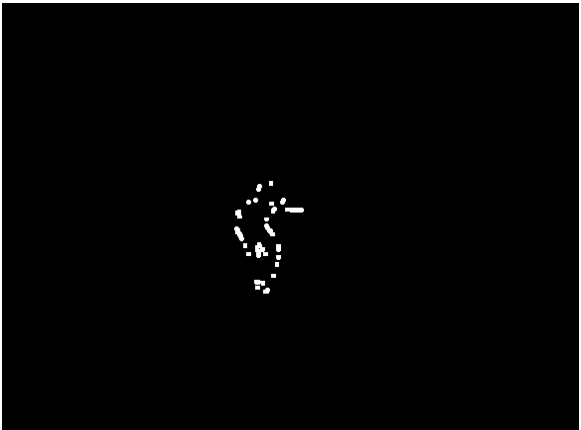

**Figure 7 (d): Dilated Image**

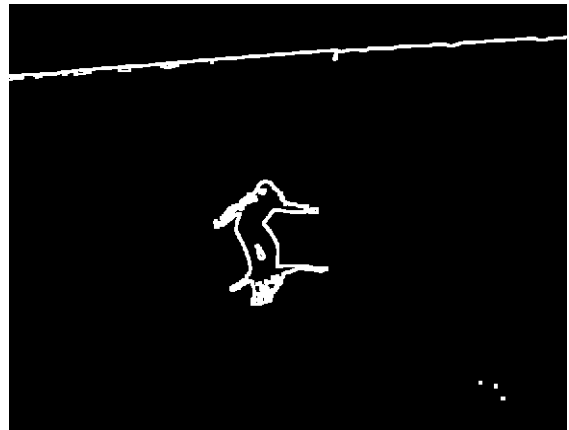

**Figure 8 (d): Dilated Image**

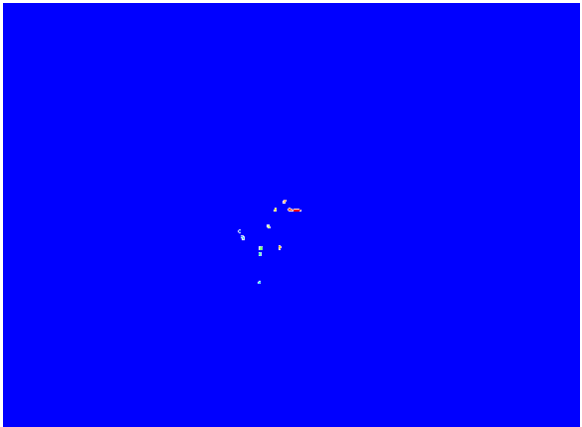

**Figure 7 (e): Watershed Transform**

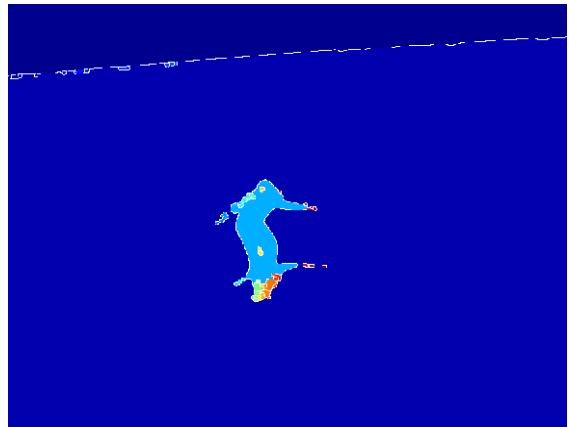

**Figure 8 (e): Watershed Transform**

## 5. CONCLUSION

The proposed image segmentation algorithm is time-efficient, i.e., having an execution time of 1.7259 seconds for segmentation with thresholding on Intel Core 2 Duo 2.2 GHz Processor. It can thus be extended to real-time segmentation.

The comparison clearly shows that thresholding is important in image segmentation. It retains most of the image details with minimum information loss. Watershed transform is an efficient segmentation tool when combined with various methods as shown. The proposed method was successfully able to segment the image effectively.